%% file: sample-sigconf.tex
\documentclass[sigconf]{acmart}

\usepackage{booktabs} 
\usepackage{dirtytalk}
\usepackage{multirow}
\usepackage{subcaption}
\usepackage{amsmath}

\acmPrice{15.00}

\acmDOI{10.1145/3274895.3274969}

\acmISBN{978-1-4503-5889-7/18/11}

\acmConference[SIGSPATIAL'18]{ACM conference}{November 2018}{Seattle, Washington USA}
\acmYear{2018}
\copyrightyear{2018}
\setcopyright{acmcopyright}

\begin{document}

\title{ What Is It Like Down There? Generating Dense Ground-Level Views and Image Features From Overhead Imagery Using Conditional Generative Adversarial Networks}


\author{Xueqing Deng}
\affiliation{%
  \institution{University of California, Merced}
}
\email{xdeng7@ucmerced.edu}

\author{Yi Zhu}
\affiliation{%
  \institution{University of California, Merced}
}
\email{yzhu25@ucmerced.edu}

\author{Shawn Newsam}
\affiliation{%
  \institution{University of California, Merced}
}
\email{snewsam@ucmerced.edu}

\renewcommand{\shortauthors}{X. Deng et al.}

\begin{abstract}
This paper investigates conditional generative adversarial networks (cGANs) to overcome a fundamental limitation of using geotagged media for geographic discovery, namely its sparse and uneven spatial distribution. We train a cGAN to generate ground-level views of a location given overhead imagery. We show the ``fake'' ground-level images are natural looking and are structurally similar to the real images. More significantly, we show the generated images are representative of the locations and that the representations learned by the cGANs are informative. In particular, we show that dense feature maps generated using our framework are more effective for land-cover classification than approaches which spatially interpolate features extracted from sparse ground-level images. To our knowledge, ours is the first work to use cGANs to generate ground-level views given overhead imagery in order to explore the benefits of the learned representations.
\end{abstract}

%
%

\ccsdesc[300]{Information systems~Geographic information systems}
\ccsdesc[300]{Computing methodologies~Image representations}
\ccsdesc[300]{Computing methodologies~Neural networks}

\keywords{Computer vision, generative adversarial networks, land-cover classification, geotagged social media}

\maketitle

\input{samplebody-conf}

\bibliographystyle{ACM-Reference-Format}
\bibliography{sample-bibliography}

\end{document}

%% file: samplebody-conf.tex
\section{Introduction}

\begin{figure}
\includegraphics[width=\linewidth]{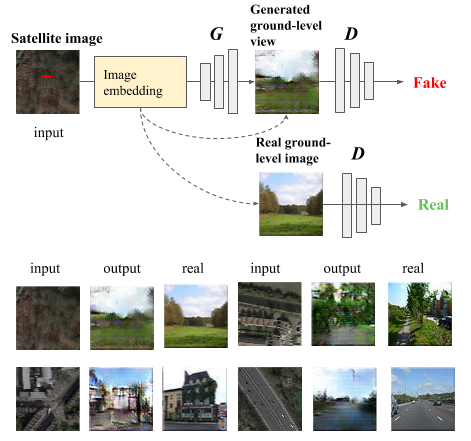}
\caption{Overview of our work and selected results. Top: Proposed conditional generative adversarial network consisting of a generator which produces ground-level views given overhead imagery, and a discriminator which helps with training the generator as well as learning useful representations. Bottom: Select overhead image patches, the ground-level views generated by our our framework, and the real ground-level images.}
\label{fig:overview}
\end{figure}

Mapping geographic phenomena on the surface of the Earth is an important scientific problem. The widespread availability of geotagged social media has enabled novel approaches to geographic discovery. In particular, \say{proximate sensing} \cite{leung2010proximate}, which uses ground-level images and videos available at sharing sites like Flickr and YouTube, provides a different perspective from remote sensing, one that can see inside buildings and detect phenomena not observable from above. Proximate sensing has been applied to map land use classes \cite{zhu2018arxiv,zhu2015land}, public sentiment \cite{zhu2016spatio}, human activity \cite{Zhu2017Activity}, air pollution \cite{li2015using}, and natural events \cite{wang2016tracking}, among other things. However, a fundamental limitation to using geotagged social media for mapping is its sparse and uneven spatial distribution. Unlike overhead imagery, it generally does not provide dense  or uniform coverage.

This limitation restricts the kinds of maps that can be generated from geotagged social media. It also presents a challenge \emph{when this data is fused with overhead imagery}. There has been great success on fusing overhead imagery with ground-level social media for problems such as land use classification \cite{liu2017ijgis_landuse,Hu2016MappingLU}. However, the maps produced by these approaches are at coarser spatial scales than the overhead imagery since the social media data is not available everywhere and therefore must be aggregated, say at the building or parcel level for land use classification. There has been less success on fusing the overhead imagery and ground-level social media at the resolution of the imagery. This is critical for producing finer scale, uniform maps as well as for areas where the aggregation units are not known a priori. 

To our knowledge, Workman et al. \cite{Workman2017AUM} were the first to fuse overhead and ground-level imagery at the spatial resolution of the overhead imagery. They combine satellite and Google Street View (GSV) imagery to predict land use, building age, etc. They overcome the nonuniform spatial distribution of the GSV imagery by using Gaussian kernels to spatially interpolate features extracted from individual GSV images to match the resolution and coverage of the satellite imagery. However, as we showed in our previous work \cite{deng2018icip}, such an approach is problematic in that it fails to infer the discontinuous spatial distribution of the ground-level image features. This motivates our novel work in this paper on more accurately estimating the spatial distribution of the \emph{ground-level image features}. We do this by asking the question of what the ground-level view looks like on a dense spatial grid in order to derive the features. While this might seem to be an intractable problem, we believe that the recently proposed conditional generative adversarial networks provide an interesting solution.

Generative adversarial networks (GANs) have shown remarkable success in generating ``fake'' images that nonetheless look realistic. The key is learning the distribution of real-looking images in the space of all possible images. GANs accomplish this implicitly through a two player game in which a generator, given random noise as input, learns to generate images which a discriminator cannot tell apart from real images. Once trained, the generator can be used to produce novel images given random noise as input.

Our problem is a bit different, though. We do not want to simply generate ground-level views that look realistic (or, more accurately, whose features are from the distribution of real images). We also want to know how these ground-level views vary spatially. We thus turn to conditional GANs (cGANs) in which the generator and discriminator are conditioned on some additional information. This auxiliary information can be a simple class label, for example when generating real-looking images of the handwritten digits 0-9, or more complex data, possibly even in a different modality \cite{mirza2014conditional}. cGANs have been used to generate photo-realistic pictures from text descriptions \cite{zhang2017stackgan} and to transfer styles between different visual domains, such as rendering a photograph as a painting or a satellite image as a map \cite{pix2pix2017}. We perform a novel investigation into using cGANs to generate ground-level views conditioned on overhead imagery in order to produce dense feature maps.

The contributions of our work are as follows:
(1) We propose a novel cGAN for generating ground-level views given overhead imagery.
(2) We explore different representations/embeddings of the overhead imagery including image patches and convolutional neural network (CNN) features.  
(3) We demonstrate that our cGAN learns informative features that can be used, for example, to perform land-cover classification.
(4) Finally, we compare the dense feature maps produced by our framework to those produced through interpolation in the context of land-cover mapping.

Our paper is organized as follows. Section 2 presents related work and Section 3 provides the technical details of our framework. Section 4 describes the datasets, provides visualizations of the generated ground-level views, and presents quantitative results of the land-cover classification. Section 5 concludes.

\section{Related work}
\subsection{Conditional Generative Adversarial Nets}
In the last few years, GANs have been explored for various computer vision problems and have shown great promise for generating detailed images \cite{goodfellow2014generative} compared to previous generative methods which resulted in smoothed images. They are still difficult to work with, however, and instability in training makes it challenging to produce high-resolution or high-quality images. A number of techniques have therefore been proposed \cite{goodfellow2017nipstutorial,radford2015unsupervised,salimans2016improved,berthelot2017began,mao2016lsgan,arjovsky2017wgan,2016arXivimplicit,Denton2015lpgan} to stabilize the training as well as improve the results.

A number of interesting applications have been studied, particularly in the conditional setting, and include semantic face completion, which aims to recover a masked face \cite{yeh2017semantic,li2017face}, image style transfer \cite{pix2pix2017,CycleGAN2017,Bousmalis2017UnsupervisedPD}, face rotation \cite{huang2017face,2017arXivface}, pose estimation \cite{ma2017pose,2017arXivpose}, face generation \cite{li2017face}, super resolution \cite{2017Ledigsrgan}, semantic image generation \cite{Dong2017SemanticIS} and so on.

\textit{\textbf{Pix2Pix}} \cite{pix2pix2017} performs image translation by processing the input images using an auto-encoder or a U-Net for the generator. A PatchGAN architecture is used for the discriminator. The skip connections in the U-Net, in particular, allow the generator to pass and thus preserve low-level characteristics of the input image, such as layout and shape, to the output image. Our problem is different though. There is no direct structural similarity between the overhead and the ground-level images as shown in Figure \ref{fig:overview}. While we expect the overhead image to be informative about what the ground-level view looks like, ours is not a style transfer problem and so the Pix2Pix framework is not appropriate.

\begin{figure*}[htbp]
\centering
\includegraphics[width=\textwidth]{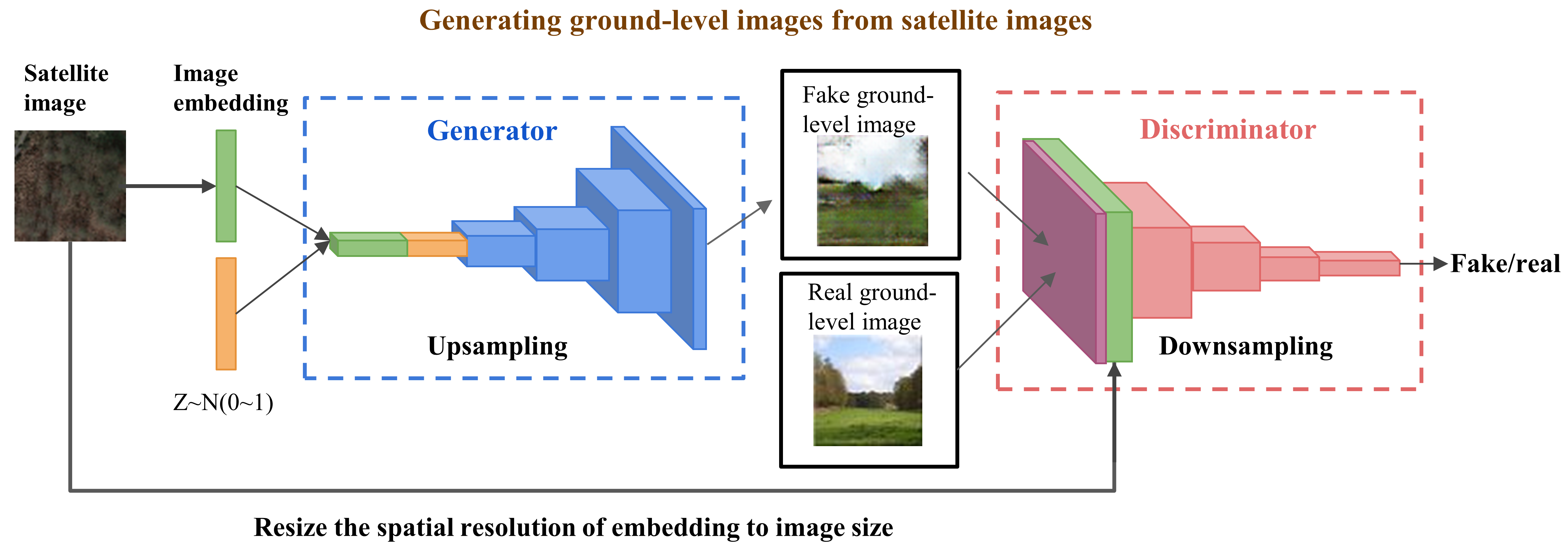}
\caption{Network architecture: The generator takes as input an overhead image patch encoded to a vector and concatenated with a random vector. Its output is a realistic-looking but ``fake'' ground-level image. The discriminator tries to tell the difference between real and fake images. These two components play a competitive game during training.}
\label{fig:pipeline}
\end{figure*}

\textit{\textbf{StackGAN}} \cite{zhang2017stackgan} proposes a two-stage GAN to transform text descriptions into photo-realistic images. Text descriptions are extracted by a pretrained encoder and then text embeddings are produced with Gaussian conditioning variables from the description in order to mitigate discontinuities which can cause instability during training. While this is not an image translation task, we take inspiration from it and investigate methods to extract embeddings of our overhead imagery.

\subsection{Representation Learning of GANs}
While GANs can amaze due to how realistic the generated images look, their real power often lies in the learned representations and the application of these representations to image analysis problems such as classification. Importantly, these representations are learned in an unsupervised manner since the training data need only consist of real images. For example, Raford et al. \cite{radford2015unsupervised} propose a deep convolutional GAN architecture (DCGAN) to not only generate photo-realistic images but also extract features for classification. A DCGAN model trained on ImageNet-1k dataset\footnote{http://www.image-net.org/} is used to extract features from the CIFAR-10 dataset\footnote{https://www.cs.toronto.edu/~kriz/cifar.html}. An SVM is then used to perform classification with respect to the CIFAR-10 classes. The DCGAN features achieve just 4\% lower accuracy than an Exemplar CNN model \cite{dosovitskiy2016discriminative} which is trained directly on the CIFAR-10 dataset. This demonstrates the ability of GANs to learn useful features in an unsupervised manner. In our experiments below, we investigate the ability of GANs to learn useful features for land-cover classification.

\subsection{Producing Dense Spatial Feature Maps}

As already mentioned, a challenge to using geotagged social media for geographic discovery is its sparse and nonuniform distribution. Researchers have therefore investigated methods to spatially interpolate the \emph{features} extracted from, for example, ground-level images to produce dense feature maps. This is the approach taken by Workman et al. in \cite{Workman2017AUM} to fuse VGG-16 features extracted from Google Street View images with high-resolution satellite imagery for dense land-use classification. However, as we showed in previous work \cite{deng2018icip}, this interpolate-then-classify approach assumes that the ground-level features vary smoothly spatially which is often not the case. Instead, in this paper, we propose a novel method for generating dense ground-level feature maps by training a cGAN to generate fake ground-level images given overhead imagery and then using the learned representations as the features. The cGAN in a sense learns to predict ``What is it like down there?'' given overhead imagery.

\subsection{Overhead to Ground-Level Cross-View Image Synthesis}
We note there has been some prior and concurrent work on overhead to ground-level cross-view image synthesis. The prior work of Zhai et al. \cite{zhai2017cvpr} learns to extract semantically meaningful features from overhead imagery by performing cross-view supervised training comparing the semantic segmentation of co-located Google Street View panoramas with segmentations generated from the overhead images that are transformed to ground-level view. The method is shown to be useful for weakly supervised overhead image segmentation, as well as a pre-training step for fully supervised segmentation; and for geo-locating and geo-orienting ground-level images. They do use the learned features to synthesize ground-level panoramas using a GAN-like architecture but these images are only visualized and not used for further analysis. And, only ground-level Street View images are considered. Street View images share significantly more structural similarity with co-located overhead images than our ground-level images which can be from any location, not just along streets.

Concurrent work by Regmi and Borji \cite{Regmi_2018_CVPR,2018arXivcrossview} uses cGANs to synthesize ground-level Google Street View images using overhead imagery. However, the work is again limited to Street View images which share more structural similarity with overhead imagery. And, the goal is to produce visually high-quality ground-level images and not to use the synthesized images or their features for further geographic analysis.

\section{Methods}

\subsection{Overview of Our Methods}
Our work has two goals. First, to generate natural-looking ground-level views given overhead imagery, and, second, to explore the learned representations for dense land-cover classification. Section \ref{sec:CGANS} briefly introduces GANs and conditional GANs, and then describes our framework for generating ground-level views given overhead imagery. Section \ref{sec:embedding} describes the different embeddings we explore to input the overhead imagery to the cGAN as well as the objective function we use to train the cGAN. Section \ref{sec:architecture} provides the network architecture and implementation details. Finally, Section \ref{sec:densefeaturemaps} describes how we access the learned representations by making modifications to the output layer of the discriminator.

\begin{table*}[!htb]
    \caption{Network architecture}
    \label{tab:arch}
    \begin{subtable}{.5\linewidth}
      \centering
        \caption{Generator}
        \begin{tabular}{ |p{1.5cm}| c |p{1cm}|p{0.8cm}|p{1cm}|p{1.6cm}| }
\hline
Name&kernel& channel in/out & In res & Out res & Input\\
\hline
deconv1&$4\times4$ & \textit{nef}+100 /1024& $1\times1$&$4\times4$ &embedding+ random noise\\
\hline
 deconv1\_bn&\multicolumn{4}{c|}{Batchnorm}&deconv1 \\
 
 \hline
 deconv2 &$4\times4$ & 1024/512 & $4\times4 $& $8\times8$ &deconv1\_bn\\

\hline
deconv2\_bn& \multicolumn{4}{c|}{Batchnorm}&deconv2 \\
 \hline
 deconv3 &$4\times4$ &512/256& $8\times8 $& $16\times16$ &deconv2\_bn \\

\hline
deconv3\_bn &\multicolumn{4}{c|}{Batchnorm}&deconv3 \\
 \hline
 deconv4 &$4\times4$ &256/128 & $16\times16 $& $32\times32$ &deconv3\_bn \\
\hline
decon4\_bn &\multicolumn{4}{c|}{Batchnorm}&deconv4 \\
 \hline
 deconv5 &$4\times4$ &128/3 & $32\times32 $& $64\times64$ &deconv4\_bn \\
\hline

\end{tabular}
    \end{subtable}%
    \begin{subtable}{.5\linewidth}
      \centering
        \caption{Discriminator}
        \begin{tabular}{ | p{1.25cm}| c |p{1cm}|p{0.8cm}|p{1cm}|p{1.6cm} | }
\hline
Name&kernel& channel in/out & In res & Out res & Input\\
\hline
conv1\_1&$4\times4$ & 3/64& $64\times64$&$32\times32$ &image\\
\hline
 conv1\_bn1&\multicolumn{4}{c|}{Batchnorm}&conv1\_1 \\
 \hline
 conv1\_2 &$4\times4$ & $nef$/64& $64\times64 $& $32\times32$ &embedding \\
\hline
 conv1\_bn2&\multicolumn{4}{c|}{Batchnorm}&conv1\_2 \\
 \hline
 conv2 &$4\times4$ & 128/256 & $32\times32 $& $16\times16$ &conv1\_bn1+ conv1\_bn2 \\

\hline
conv2\_bn& \multicolumn{4}{c|}{Batchnorm}&conv2 \\
 \hline
 conv3 &$4\times4$ &125/512& $16\times16 $& $8\times8$ &conv2\_bn \\

\hline
conv3\_bn &\multicolumn{4}{c|}{Batchnorm}&conv3 \\
 \hline
 conv4 &$4\times4$ &512/1024 & $8\times8 $& $4\times4$ &conv3\_bn \\
\hline
con4\_bn &\multicolumn{4}{c|}{Batchnorm}&conv4 \\
 \hline
 conv5 &$4\times4$ &1024/1 & $4\times4 $& $1\times1$ &conv4\_bn \\
\hline

\end{tabular}
    \end{subtable} 
\end{table*}

\subsection{GANs and cGANs}
\label{sec:CGANS}
GANs \cite{goodfellow2014generative} consist of two components, a generator and a discriminator. As  shown in Figure \ref{fig:pipeline}, the generator $G$ generates realistic looking but fake images by upsampling vectors of random noise. The discriminator's goal is to distinguish between real and fake images through downsampling. GANs learn generative models through adversarial training. That is, $G$ tries to fool $D$. $G$ and $D$ are trained simultaneously. $G$ is optimized to reproduce the true data distribution by generating images that are difficult to distinguish from real images by $D$. Meanwhile, $D$ is optimized to differentiate fake images generated by $G$ from real images. Overall, the training procedure is similar to a two-player min-max game with the following objective function

\begin{equation}
\begin{split}
\min \limits_{G}\max \limits_{D} V(D,G)=\mathbb{E}_{x\backsim p_{data}(x)}[\log D(x)] + \\
\mathbb{E}_{z\backsim p_{data}(z)}[1-  \log D(G(z))]
\end{split}
\end{equation}
where $z$ is the random noise vector and $x$ is the real image. Once trained, $G$ can be used to generate new, unseen images given random vectors as input. In practice, rather than train $G$ to minimize
$1- \log D(G(z))$, we train $G$ to maximize $\log D(G(z))$, as demonstrated in \cite{goodfellow2014generative}.

GANs, as a generative model, learn the distribution of the entire training dataset. Conditional GANs were therefore introduced to learn distributions conditioned on some auxiliary information. For example, a GAN can be trained to generate realistic-looking images of hand written digits from the MNIST dataset. However, if we want images of a specific digit, a ``1'' for example, the generative model needs to be conditioned on this information. In cGANs, the auxiliary information $y$ is incorporated through hidden layers separately in both the generator and discriminator. $y$ can take many forms, such as class labels \cite{mirza2014conditional}, text embeddings for generating images from text \cite{pmlr-v48-reed16,zhang2017stackgan}, and images for image translation \cite{pix2pix2017,CycleGAN2017}. The cGAN objective function becomes

\begin{equation}
\begin{split}
\min \limits_{G}\max \limits_{D} V(D,G)=\mathbb{E}_{x\backsim p_{data}(x)}[\log D(x| y)] + \\
\mathbb{E}_{z\backsim p_{data}(z)}[1-  \log D(G(z|y))].
\end{split}
\end{equation}

In the context of our problem, we use overhead imagery as the auxiliary information to generate natural-looking ground-level views. However, ours is not an image translation problem since there is no direct structural similarity between the overhead images and the ground-level views. We therefore consider embeddings of the overhead imagery.

\subsection{Overhead Image Embedding}
\label{sec:embedding}

The overarching premise of our novel framework is that overhead imagery contains information about what things look like on the ground. It is not obvious, however, how to represent this information so that it can be input to a cGAN to generate ground-level views. We know that simply feeding a 2D overhead image patch to the cGAN and then performing image translation does not make sense due to the lack of structural similarity between the overhead and ground-level images. We are also constrained by the fact that the input to GANs/cGANs cannot be too large otherwise the learning will become unstable.

Below, we describe three types of embeddings, each of which produces a 1D vector which is concatenated with the noise vector and then input to the generator. As is standard in cGANs, the overhead image information is also provided to the discriminator. See Figure \ref{fig:pipeline} for details.

The objective function for training our cGAN to generate ground-level views from overhead imagery becomes

\begin{equation}
\begin{split}
&L_D=\mathbb{E}_{(x,\varphi(I_s))\backsim p_{data}(I_g,\varphi(I_s))}+ \\
&\mathbb{E}_{z\backsim p_{z},\varphi(I_s)\backsim p_{data}}[1-  \log D(G(z,\varphi(I_s)),\varphi(I_s))]
\end{split}
\end{equation}

\begin{equation}
L_G=\mathbb{E}_{z\backsim p_{z},\varphi(I_s)\backsim p_{data}}[1-  \log D(G(z,\varphi(I_s)),\varphi(I_s))]
\end{equation}
where $I_s$ and $I_g$ refer to satellite image and ground-level images respectively, and $\varphi(I)$ is the overhead image embedding.

\textbf{\textit{Grayscale image patch embedding}} Our baseline embedding extracts a $10\times10$ pixel patch from the overhead imagery centered on where we want to generate the ground-level view, computes the grayscale pixel values as the average RGB values, reshapes the patch into a vector, and normalizes the components to range between -1 and 1. This results in a 100D vector.

\textbf{\textit{HSV image patch embedding}} In order to investigate whether the color information in the overhead imagery is useful for our problem, we convert the $10\times10$ patches from RGB to HSV colorspaces and then form a normalized, 300D vector.

\begin{figure*}[htbp]
\includegraphics[width=\textwidth]{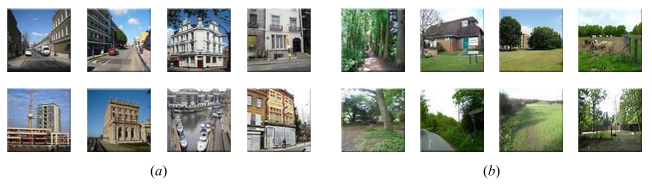}
\caption{Examples of \textbf{Geograph} images for (a) urban locations and (b) rural locations.}
\label{fig:examples_geograph}
\end{figure*}

\textbf{\textit{CNN image feature embedding}} The above embeddings are straightforward encodings of the overhead image patches. We also consider using a pre-trained VGG-16 model to encode the image patches as CNN features. This results in 1024D vectors which we reduce to 25D through dimensionality reduction. The motivation here is that CNNs have been shown to produce effective image encodings for a range of computer vision tasks.

\subsection{Network Architecture and Implementation Details}
\label{sec:architecture}
GANs architectures can consist of either multi-layer perceptrons or convolutional networks. We adopt CNNs as they have been shown to be more stable during training \cite{radford2015unsupervised}. Our CNNs consist of convolutional (conv) layers, followed by batchnorm \cite{ioffe2016batchnorm} and leakyReLU steps, and so are referred to as a conv-batchnorm-leakyReLU architecture. Strided convolutions are used to increase or decrease the image resolution instead of maxpooling. Table \ref{tab:arch} provides the details of our generator and discriminator networks. In Table \ref{tab:arch}, (a) and (b) are the network architectures for generator and discriminator individually, where $deconv$ denotes a transposed convolution layer, $conv$ denotes a convolution layer, $bn$ denotes batchnorm and $nef$ denotes the number of dimensions of the embedding function output.

The input to our generator is the concatenation of a 100D random noise vector and an $nef$ dimensional embedding of the overhead image patch. We concatenate these vectors before inputting them to the conv-batchnorm-leakyReLU architecture similar to the work on text-to-image generation in \cite{zhang2017stackgan}. This is different from the work in \cite{mirza2014conditional} in which the two vectors are separately fed through convolutional layers and later concatenated. We find that concatenating first makes the training more stable. The output of our generator is a $64\times64$ pixel RGB ground-level image.

Our discriminator follows the same conv-batchnorm-leakyReLU architecture but the last layer is a sigmoid activation function which outputs a binary value corresponding to real or fake. The discriminator takes as input a ground-level image, real or fake, and the auxiliary information in the form of the overhead image patch. In the case of a real image, the overhead image patch is the actual overhead view of where the image is located. In the case of a fake image, the overhead image patch is what was used by the generator to produce the image. The output of the discriminator is its belief of whether the ground-level image is real or fake. Two different losses are used to train the discriminator depending on whether the input ground-level image is real or fake.

We implement our deep learning framework using PYTORCH\footnote{https://pytorch.org} and ADAM \cite{Kingma2014AdamAM} is used as our optimizer. The initial learning rate is set to 0.0002. We train our cGAN for 400 epochs with a batch size of 128 on one NVIDIA GTX 980Ti GPU.

\subsection{Generating Dense Feature Maps For Land-Cover Classification}
\label{sec:densefeaturemaps}
Our goal is not just to see how well our generator can produce ground-level views given overhead imagery, but also to investigate whether our entire cGAN, the generator and discriminator, can learn novel representations for tasks such as classification. As an example application, we explore whether these representations are effective for land-cover classification. This would then allow us to generate dense ground-level feature maps given only overhead imagery.

We modify our cGAN to output features as follows. Following \cite{radford2015unsupervised}, we remove the last, sigmoid layer of our discriminator and add an average pooling layer. Our cGAN then outputs a 1024D feature vector given an overhead image patch. This feature can then be used to perform analysis at the spatial resolution of the overhead imagery (by using overlapping patches), such as per pixel classification, image segmentation, etc.

\begin{figure*}[h]
\centering
\includegraphics[width=\linewidth]{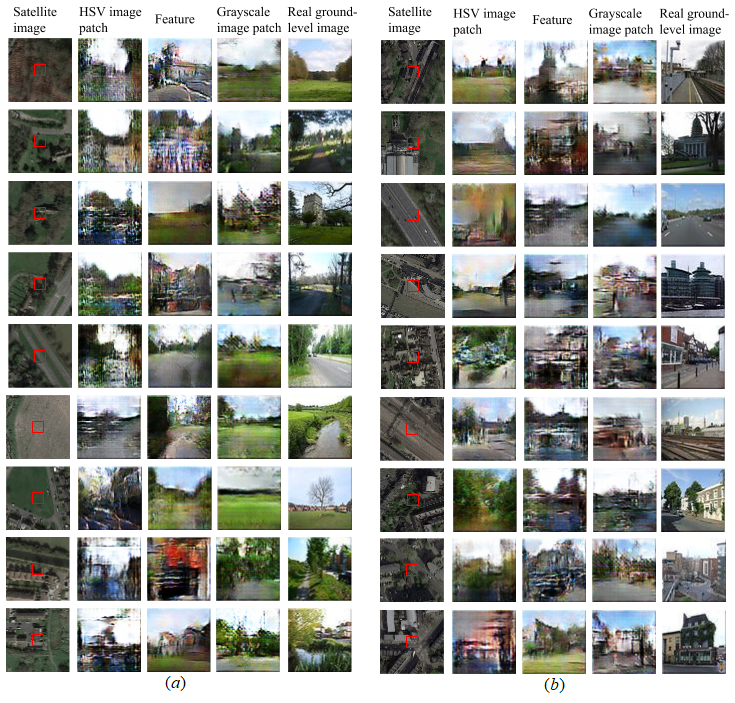}
\caption{Selected images generated by our proposed cGAN on (a) rural locations and (b) urban locations. The columns for each group from left to right are the input overhead images (extracted image patches are indicated by the red boxes), the cGAN generated images generated using HSV, CNN, and grayscale encoded patches, and the real ground-level images. }
\label{fig:GAN_imgs}
\end{figure*}

\begin{table*}
\caption{Land-cover classification accuracy with features}
\label{tab:svm-acc}
\centering
\begin{tabular}{ c| c |c|c|c }
\hline
\multirow{2}{4em}{Classifier}& \multicolumn{2}{c|}{Features}&\multirow{2}{5em}{Dimension}&\multirow{2}{4em}{Accuracy}\\
\cline{2-3}
 &Type&Name & & \\
\hline
\multirow{12}{4em}{SVM} &\multirow{4}{18em}{Embedding from Google satellite images} & Grayscale image patches&100 & 93.17 \\  
 &  &HSV image patches &300 &  \textbf{93.47}\\
& &VGG-extracted image features & 25&93.19 \\
\cline{2-5}
&\multirow{4}{18em}{cGAN generated image features conditioned on different embeddings} &Grayscale image patches&1024 & \textbf{82.33} \\
 & &HSV image patches feature &1024& 74.33 \\
 & &VGG-extracted image features& 1024 & 61.82\\ 
\cline{2-5}
  &\multirow{4}{18em}{cGAN generated image features conditioned on different embeddings + embeddings} &  Grayscale image patches + image patches& 1024+100 &\textbf{86.34} \\
&& HSV image patches + image patches &1024+300& 75.13 \\
& &VGG-extracted image features + image features & 1024+25& 65.94  \\
 \hline
 ResNet-34&Ground-level images& Interpolated CNN features & 512 & 58.5\\
 \hline
\end{tabular}
\end{table*}

\section{Experiments}

\subsection{Datasets and Land-Cover Classes}
Our framework requires co-located ground-level and overhead imagery. We download ground-level images with known location from the Geograph API\footnote{http://www.geograph.org.uk/}. We download georeferenced overhead imagery from the Google Map Static API\footnote{https://developers.google.com/maps/documentation/maps-static/intro}.

For land-cover classification, we use the ground-truth land-cover map LCM2015\footnote{https://eip.ceh.ac.uk/lcm/lcmdata} to construct our training and test datasets. This map provides land-cover classes for the entire United Kingdom on a 1km grid. We group these classes into two super-classes, urban and rural, and limit our study to a 71km$\times$71km region containing London. The Geograph images are labeled as urban or rural based on the LCM2015 label of the 1km$\times$1km grid cell from which they are downloaded. We realize this label propagation likely results in some noisy labels in our dataset.

Figure \ref{fig:examples_geograph} shows that the Geograph images corresponding to our two land-cover classes are visually very different.

\subsection{Preprocessing Data}
We resize our real ground-level images to measure $64\times64$ pixels to match the size of the images produced by our generator. This allows us to compare classification results of using the real images with the results of using the fake images.

\subsection{Generating Ground-Level Views Given Overhead Imagery}
We trained our cGAN using 4,000 Geograph images and co-located overhead image patches split evenly between urban and rural locations. We then used our generator to produce ground-level views given overhead image patches at other locations. Figure \ref{fig:GAN_imgs} shows the overhead image patches and generated ground-level views corresponding to the three different overhead image embeddings. Also shown are the true ground-level images.

Despite the difficulty of our task, the results in Figure \ref{fig:GAN_imgs} demonstrate that our proposed cGAN framework is able to produce surprisingly reasonable ground-level views given the overhead image patches. The results of the grayscale embedding are the most realistic looking and generally align well with the real images. This is particularly true for the rural scenes with lots of grass, trees and sky. Urban scenes are more complex with detailed objects with crisp boundaries and the results look less realistic. However, similar to the real images, \emph{the generated ground-level views corresponding to the two land-cover classes are visually very distinct}, at least for the grayscale embedding.

The results of the HSV embedding are not as realistic looking as that of the grayscale embedding. This is somewhat surprising since it seems the additional color information would be informative. The 300D vector embeddings might simply be too large for training the cGAN.

The results of the CNN feature embedding are poor. The images are not realistic looking and there is not much distinction between the two land-cover classes. We will investigate this further in future work as it seems the CNN features should be able to encode the overhead image information.

In summary, while it is unlikely the generated ground-level views would be mistaken for real images, the results of the grayscale and HSV embeddings are very representative of what things look like on the ground. We believe this is an impressive outcome given that these images are generated solely using $10\times10$ pixel overhead image patches.


\subsection{Generating Ground-Level Image Features Given Overhead Imagery}
\label{sec:GL_image_features}
We now investigate whether our trained cGAN has learned representations useful for other tasks such as classification. We compare land-cover classification using our learned representation with interpolating between real ground-level images.

We download Geograph images and co-located overhead imagery corresponding to 20,000 locations. The land-cover classes of these locations are known from the LCM2015 ground truth map. These locations are split into 16,000 training locations and 4,000 test locations.

We first make sure that the overhead image patches themselves are representative of the land-cover classes. We train SVM classifiers (with RBF kernels) using the different embeddings of the overhead image patches as input (again, we know the classes of these patches). We then apply the SVMs to the overhead image patches corresponding to the test locations. The results are shown at the top of Table \ref{tab:svm-acc}. All three classifiers achieve over 93\% accuracy on the test set which indicates the overhead image patches are representative of the two classes and are thus suitable candidates as auxiliary information for our cGANs.

We now evaluate our learned representation. For each of the 20,000 locations, we generate a 1024D feature by applying our modified cGAN (see Section \ref{sec:densefeaturemaps}) to the overhead image patch for that location. We then train and test an SVM classifier. This is done separately for the three different overhead image embeddings. The classification results are shown in Table \ref{tab:svm-acc}. The grayscale embedding performs the best, achieving an accuracy of 82.33\%, demonstrating that our cGAN has learned a useful representation. For context, learning a ResNet-34 \cite{he2016deep} classifier on the 16,000 real ground-level training images and then applying it to the 4,000 real ground-level test images results in 88.2\% accuracy. The performance of our learned representation is not far off this gold standard.

The classification results of the HSV and CNN feature embeddings are 72.12\% and 61.82\% respectfully. Again, the grayscale embedding seems best.

We now evaluate the performance of interpolating between real ground-level images. We divide the ResNet-34 classifier learned from the 16,000 real ground-level training images into a feature extractor and a classifier. The feature extractor is applied to the 16,000 training images to derive a 512D feature for each of the 16,000 locations. The 512D features at the 4,000 test locations are derived by spatially interpolating the features at the 16,000 training locations using a Gaussian kernel. The classifier is then applied to the interpolated features. As shown in Table \ref{tab:svm-acc}, this results in an accuracy of only 58.5\%, much worse than our proposed cGAN framework which is able to achieve 82.33\%.

\begin{figure*}
\centering
\includegraphics[width=\textwidth]{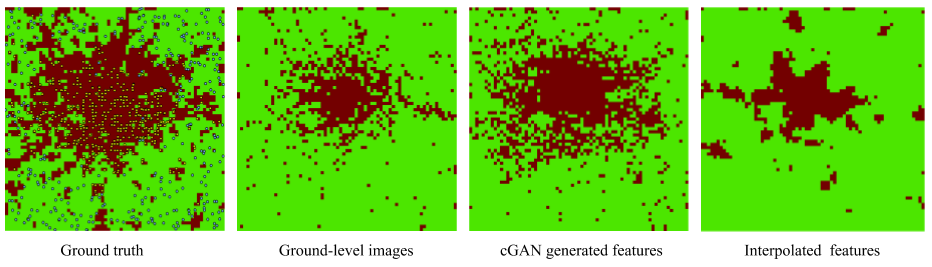}
\caption{Ground truth and predicted land-cover maps. From left to right: ground truth map; map generated using ResNet-34 applied to a dense sampling of ground-level images; map generated using cGAN generated features; and map generated using interpolated features. (brown: urban, green: rural) The dots in the ground truth represent the locations of the sparse ground-level images used in the interpolation (yellow: urban, blue: rural)}
\label{fig:lcm}
\end{figure*}

Finally, for completeness, we concatenate the cGAN learned representations and the overhead image embeddings, and train and evaluate an SVM classifier. As shown in Table \ref{tab:svm-acc}, this improves upon the performance of only using the cGAN representations.

\subsection{Creating Dense Feature Maps}

\begin{table}
\caption{Land-cover classification with dense feature map}
\label{tab:lcc_quant}
\begin{tabular}{ c| c }
\hline
Name&Accuracy \\
\hline
Geograph images &69.07 \\
cGAN generated features & \textbf{73.14} \\
Interpolated features & 65.86\\
\hline
\end{tabular}
\end{table}

In our final experiment, we use our cGAN framework to produce a dense ground-level feature map given overhead imagery. This feature map is then used to generate a land-cover map. We compare this map with ones produced by classifying densely-sampled ground-level images as well by an interpolate-then-classify \cite{Workman2017AUM,deng2018icip} approach applied to sparsely-sampled images.

The left image in Figure \ref{fig:lcm} shows the ground truth map based on LCM2015. Again, this is a 71$\times$71 gridded region in which each cell corresponds to a 1km$\times$1km square.

To produce a land-cover map based on densely-sampled ground-level images, we download 10 Geograph images for each cell, keeping track of the locations. This results in $71\times71\times10=50,410$ images in total. The ResNet-34 classifier trained earlier is applied to the 10 images in each grid cell and the majority label is assigned to the cell. This results in the second map from the left in Figure \ref{fig:lcm}.

To produce a land-cover map based on our cGAN framework, we use our modified cGAN  to extract features from overhead patches at the locations of the 50,410 images. We then apply the SVM classifier trained earlier to the features for the 10 locations in each grid cell and again assign the majority label to the cell. This results in the third map from the left in Figure \ref{fig:lcm}.

Finally, to produce a land-cover map using an interpolate-then-classify approach, we use the ResNet-34 feature extractor to extract features from just 836 of the 50,410 ground-level images. Spatial interpolation is then used to estimate the features at the remaining locations and the ResNet-34 classifier is applied. Implementation details can be found in \cite{deng2018icip}. The majority label is again used to label the grid cells. This results in the map on the right in Figure \ref{fig:lcm}.

Qualitatively, the map produced using the proposed cGAN framework is more similar to the ground truth than those generated using the densely-sampled ground-level images and the interpolate-then-classify approach. The ground-level images are quite heterogeneous and can vary quite a bit within a 1km$\times$1km region. A cell will be labeled incorrectly if it happens that the majority of the images depict a class other than the ground truth. And, as expected, the interpolate-then-classify approach results in a smoothed map. Interpolation is not able to detect islands of one class that are surrounded by another class.

Table \ref{tab:lcc_quant} shows a quantitative comparison between the ground truth and predicted maps. Overall accuracy is computed as the percentage of grid cells for which the ground truth and predicted maps agree. The proposed cGAN framework is shown to perform the best.

\begin{table*}[htbp]
\caption{Land-cover classification accuracy using ground-level images}
\label{tab:resnet-acc}
\centering
\begin{tabular}{ c|c |c }
\hline
Classifier&Images&Accuracy\\
\hline
\multirow{5}{4.5em}{ResNet-34} & Real ground-level images & 88.2\\ 
 & cGAN generated fake images with grayscale overhead image embedding & 62.8\\
& cGAN generated fake images with HSV overhead image embedding & 62.5\\
& cGAN generated fake images with CNN features overhead image embedding & 61.8\\

 \hline
\end{tabular}
\end{table*}

\subsection{Limitations of the Generated Views}
We investigate whether the generated images themselves, as opposed to the learned representations, are useful for classification. We generate 20,000 fake images corresponding to the 20,000 locations used in Section \ref{sec:GL_image_features} using overhead image patches. We then learn a ResNet-34 classifier using the 16,000 fake images corresponding to the training locations and apply it to the 4,000 fake images corresponding to the test locations. The classification accuracies corresponding to the three overhead image embeddings are shown in Table \ref{tab:resnet-acc}. Also shown on the top row is the accuracy of the ResNet-34 classifier learned and applied to the real ground-level images. The fake images are seen not to be as effective for classification as the real images or the learned representations (see Table \ref{tab:svm-acc}). This is likely a result of our generated images lacking the details that the ResNet-34 classifier is able to exploit in real images.

\section{Conclusions and future work}
We investigate cGANs for generating ground-level views and the corresponding image features given overhead imagery. The generated ground-level images look natural, although, as expected, they lack the details of real images. They do capture the visual distinction between urban and rural scenes. We show the learned representations are effective as image features for land-cover classification. In particular, the representations are almost as effective at classifying locations into urban and rural classes as the real ground-level images. We use the cGANs to generate dense feature maps. These feature maps are more effective for producing land-cover maps than dense samples of ground-level images. Our proposed method is not limited to land-cover classification. It provides a framework to create dense feature maps for other applications. 

This represents our preliminary work on this problem. We plan to develop cGANs that can generate more detailed ground-level views that can be used directly for image classification, etc. The training of the cGANs is still very unstable. We will therefore also investigate other techniques and architectures to make the training of cGANS more stable for our particular problem. 

\section{Acknowledgments}
This work was funded in part by a National Science Foundation CAREER grant, \#IIS-1150115. We gratefully acknowledge the
support of NVIDIA Corporation through the donation of the GPU card used in this work

%% file: sample-sigconf.bbl

\begin{thebibliography}{41}


\ifx \showCODEN    \undefined \def \showCODEN     #1{\unskip}     \fi
\ifx \showDOI      \undefined \def \showDOI       #1{#1}\fi
\ifx \showISBNx    \undefined \def \showISBNx     #1{\unskip}     \fi
\ifx \showISBNxiii \undefined \def \showISBNxiii  #1{\unskip}     \fi
\ifx \showISSN     \undefined \def \showISSN      #1{\unskip}     \fi
\ifx \showLCCN     \undefined \def \showLCCN      #1{\unskip}     \fi
\ifx \shownote     \undefined \def \shownote      #1{#1}          \fi
\ifx \showarticletitle \undefined \def \showarticletitle #1{#1}   \fi
\ifx \showURL      \undefined \def \showURL       {\relax}        \fi
\providecommand\bibfield[2]{#2}
\providecommand\bibinfo[2]{#2}
\providecommand\natexlab[1]{#1}
\providecommand\showeprint[2][]{arXiv:#2}

\bibitem[\protect\citeauthoryear{Arjovsky, Chintala, and Bottou}{Arjovsky
  et~al\mbox{.}}{2017}]%
        {arjovsky2017wgan}
\bibfield{author}{\bibinfo{person}{M. Arjovsky}, \bibinfo{person}{S. Chintala},
  {and} \bibinfo{person}{L. Bottou}.} \bibinfo{year}{2017}\natexlab{}.
\newblock \showarticletitle{{W}asserstein Generative Adversarial Networks}. In
  \bibinfo{booktitle}{\emph{Proceedings of the 34th International Conference on
  Machine Learning}}, \bibfield{editor}{\bibinfo{person}{Doina Precup} {and}
  \bibinfo{person}{Yee~Whye Teh}} (Eds.), Vol.~\bibinfo{volume}{70}.
  \bibinfo{publisher}{PMLR}, \bibinfo{address}{International Convention Centre,
  Sydney, Australia}, \bibinfo{pages}{214--223}.
\newblock


\bibitem[\protect\citeauthoryear{{Berthelot}, {Schumm}, and {Metz}}{{Berthelot}
  et~al\mbox{.}}{2017}]%
        {berthelot2017began}
\bibfield{author}{\bibinfo{person}{D. {Berthelot}}, \bibinfo{person}{T.
  {Schumm}}, {and} \bibinfo{person}{L. {Metz}}.}
  \bibinfo{year}{2017}\natexlab{}.
\newblock \showarticletitle{{BEGAN: Boundary Equilibrium Generative Adversarial
  Networks}}.
\newblock \bibinfo{journal}{\emph{ArXiv e-prints}} (\bibinfo{date}{March}
  \bibinfo{year}{2017}).
\newblock
\showeprint[arxiv]{cs.LG/1703.10717}


\bibitem[\protect\citeauthoryear{Bousmalis, Silberman, Dohan, Erhan, and
  Krishnan}{Bousmalis et~al\mbox{.}}{2017}]%
        {Bousmalis2017UnsupervisedPD}
\bibfield{author}{\bibinfo{person}{K. Bousmalis}, \bibinfo{person}{N.
  Silberman}, \bibinfo{person}{D. Dohan}, \bibinfo{person}{D. Erhan}, {and}
  \bibinfo{person}{D.~Krishnan Krishnan}.} \bibinfo{year}{2017}\natexlab{}.
\newblock \showarticletitle{Unsupervised Pixel-Level Domain Adaptation with
  Generative Adversarial Networks}. In \bibinfo{booktitle}{\emph{Proceedings of
  the IEEE Conference on Computer Vision and Pattern Recognition}}.
  \bibinfo{pages}{95--104}.
\newblock


\bibitem[\protect\citeauthoryear{{Chen}, {Shen}, {Wei}, {Liu}, and
  {Yang}}{{Chen} et~al\mbox{.}}{2017}]%
        {2017arXivpose}
\bibfield{author}{\bibinfo{person}{Y. {Chen}}, \bibinfo{person}{C. {Shen}},
  \bibinfo{person}{X.-S. {Wei}}, \bibinfo{person}{L. {Liu}}, {and}
  \bibinfo{person}{J. {Yang}}.} \bibinfo{year}{2017}\natexlab{}.
\newblock \showarticletitle{{Adversarial PoseNet: A Structure-aware
  Convolutional Network for Human Pose Estimation}}.
\newblock \bibinfo{journal}{\emph{ArXiv e-prints}} (\bibinfo{date}{April}
  \bibinfo{year}{2017}).
\newblock
\showeprint[arxiv]{cs.CV/1705.00389}


\bibitem[\protect\citeauthoryear{Deng, Zhu, and Newsam}{Deng
  et~al\mbox{.}}{2018}]%
        {deng2018icip}
\bibfield{author}{\bibinfo{person}{X. Deng}, \bibinfo{person}{Y. Zhu}, {and}
  \bibinfo{person}{S. Newsam}.} \bibinfo{year}{2018}\natexlab{}.
\newblock \showarticletitle{Spatial Morphing Kernel Regression for Feature
  Interpolation}. In \bibinfo{booktitle}{\emph{25th IEEE International
  Conference on Image Processing (ICIP)}}. \bibinfo{pages}{2182--2186}.
\newblock


\bibitem[\protect\citeauthoryear{Denton, Chintala, Szlam, and Fergus}{Denton
  et~al\mbox{.}}{2015}]%
        {Denton2015lpgan}
\bibfield{author}{\bibinfo{person}{E. Denton}, \bibinfo{person}{S. Chintala},
  \bibinfo{person}{A. Szlam}, {and} \bibinfo{person}{R. Fergus}.}
  \bibinfo{year}{2015}\natexlab{}.
\newblock \showarticletitle{Deep Generative Image Models Using a Laplacian
  Pyramid of Adversarial Networks}. In \bibinfo{booktitle}{\emph{Proceedings of
  28th International Conference on Neural Information Processing Systems}}.
  \bibinfo{pages}{1486--1494}.
\newblock


\bibitem[\protect\citeauthoryear{Dong, Yu, Wu, and Guo}{Dong
  et~al\mbox{.}}{2017}]%
        {Dong2017SemanticIS}
\bibfield{author}{\bibinfo{person}{H. Dong}, \bibinfo{person}{S. Yu},
  \bibinfo{person}{C. Wu}, {and} \bibinfo{person}{Y. Guo}.}
  \bibinfo{year}{2017}\natexlab{}.
\newblock \showarticletitle{Semantic Image Synthesis via Adversarial Learning}.
  In \bibinfo{booktitle}{\emph{Proceedings of the IEEE International Conference
  on Computer Vision}}. \bibinfo{pages}{5707--5715}.
\newblock


\bibitem[\protect\citeauthoryear{Dosovitskiy, Fischer, Springenberg,
  Riedmiller, and Brox}{Dosovitskiy et~al\mbox{.}}{2016}]%
        {dosovitskiy2016discriminative}
\bibfield{author}{\bibinfo{person}{A. Dosovitskiy}, \bibinfo{person}{P.
  Fischer}, \bibinfo{person}{J. Springenberg}, \bibinfo{person}{M. Riedmiller},
  {and} \bibinfo{person}{T. Brox}.} \bibinfo{year}{2016}\natexlab{}.
\newblock \showarticletitle{Discriminative Unsupervised Feature Learning with
  Exemplar Convolutional Neural Networks}.
\newblock \bibinfo{journal}{\emph{IEEE Transactions on Pattern Analysis and
  Machine Intelligence}} \bibinfo{volume}{38}, \bibinfo{number}{9}
  (\bibinfo{year}{2016}), \bibinfo{pages}{1734--1747}.
\newblock


\bibitem[\protect\citeauthoryear{{Goodfellow}}{{Goodfellow}}{2017}]%
        {goodfellow2017nipstutorial}
\bibfield{author}{\bibinfo{person}{I. {Goodfellow}}.}
  \bibinfo{year}{2017}\natexlab{}.
\newblock \showarticletitle{{NIPS 2016 Tutorial: Generative Adversarial
  Networks}}.
\newblock \bibinfo{journal}{\emph{ArXiv e-prints}} (\bibinfo{date}{Dec.}
  \bibinfo{year}{2017}).
\newblock
\showeprint[arxiv]{cs.LG/1701.00160}


\bibitem[\protect\citeauthoryear{Goodfellow, Pouget-Abadie, Mirza, Xu,
  Warde-Farley, Ozair, Courville, and Bengio}{Goodfellow et~al\mbox{.}}{2014}]%
        {goodfellow2014generative}
\bibfield{author}{\bibinfo{person}{I. Goodfellow}, \bibinfo{person}{J.
  Pouget-Abadie}, \bibinfo{person}{M. Mirza}, \bibinfo{person}{B. Xu},
  \bibinfo{person}{D. Warde-Farley}, \bibinfo{person}{S. Ozair},
  \bibinfo{person}{A. Courville}, {and} \bibinfo{person}{Y. Bengio}.}
  \bibinfo{year}{2014}\natexlab{}.
\newblock \showarticletitle{Generative Adversarial Nets}. In
  \bibinfo{booktitle}{\emph{Advances in Neural Information Processing
  Systems}}. \bibinfo{pages}{2672--2680}.
\newblock


\bibitem[\protect\citeauthoryear{He, Zhang, Ren, and Sun}{He
  et~al\mbox{.}}{2016}]%
        {he2016deep}
\bibfield{author}{\bibinfo{person}{K. He}, \bibinfo{person}{X. Zhang},
  \bibinfo{person}{S. Ren}, {and} \bibinfo{person}{J. Sun}.}
  \bibinfo{year}{2016}\natexlab{}.
\newblock \showarticletitle{Deep Residual Learning for Image Recognition}. In
  \bibinfo{booktitle}{\emph{Proceedings of the IEEE Conference on Computer
  Vision and Pattern Recognition}}. \bibinfo{pages}{770--778}.
\newblock


\bibitem[\protect\citeauthoryear{Hu, Yang, Li, and Gong}{Hu
  et~al\mbox{.}}{2016}]%
        {Hu2016MappingLU}
\bibfield{author}{\bibinfo{person}{T. Hu}, \bibinfo{person}{J. Yang},
  \bibinfo{person}{X. Li}, {and} \bibinfo{person}{P. Gong}.}
  \bibinfo{year}{2016}\natexlab{}.
\newblock \showarticletitle{Mapping Urban Land Use by Using Landsat Images and
  Open Social Data}.
\newblock \bibinfo{journal}{\emph{Remote Sensing}}  \bibinfo{volume}{8}
  (\bibinfo{year}{2016}), \bibinfo{pages}{151}.
\newblock


\bibitem[\protect\citeauthoryear{Huang, Zhang, Li, and He}{Huang
  et~al\mbox{.}}{2017}]%
        {huang2017face}
\bibfield{author}{\bibinfo{person}{R. Huang}, \bibinfo{person}{S. Zhang},
  \bibinfo{person}{T. Li}, {and} \bibinfo{person}{R. He}.}
  \bibinfo{year}{2017}\natexlab{}.
\newblock \showarticletitle{Beyond Face Rotation: Global and Local Perception
  GAN for Photorealistic and Identity Preserving Frontal View Synthesis}. In
  \bibinfo{booktitle}{\emph{Proceedings of the IEEE International Conference on
  Computer Vision}}. \bibinfo{pages}{2458--2467}.
\newblock


\bibitem[\protect\citeauthoryear{Ioffe and Szegedy}{Ioffe and Szegedy}{2015}]%
        {ioffe2016batchnorm}
\bibfield{author}{\bibinfo{person}{S. Ioffe} {and} \bibinfo{person}{C.
  Szegedy}.} \bibinfo{year}{2015}\natexlab{}.
\newblock \showarticletitle{Batch Normalization: Accelerating Deep Network
  Training by Reducing Internal Covariate Shift}. In
  \bibinfo{booktitle}{\emph{32nd International Conference on Machine Learning}}
  \emph{(\bibinfo{series}{Proceedings of Machine Learning Research})},
  \bibfield{editor}{\bibinfo{person}{Francis Bach} {and} \bibinfo{person}{David
  Blei}} (Eds.), Vol.~\bibinfo{volume}{37}. \bibinfo{publisher}{PMLR},
  \bibinfo{address}{Lille, France}, \bibinfo{pages}{448--456}.
\newblock


\bibitem[\protect\citeauthoryear{Isola, Zhu, Zhou, and Efros}{Isola
  et~al\mbox{.}}{2017}]%
        {pix2pix2017}
\bibfield{author}{\bibinfo{person}{P. Isola}, \bibinfo{person}{J. Zhu},
  \bibinfo{person}{T. Zhou}, {and} \bibinfo{person}{A.~A Efros}.}
  \bibinfo{year}{2017}\natexlab{}.
\newblock \showarticletitle{Image-to-Image Translation with Conditional
  Adversarial Networks}. In \bibinfo{booktitle}{\emph{Proceedings of the IEEE
  International Conference on Computer Vision}}.
\newblock


\bibitem[\protect\citeauthoryear{Kingma and Ba}{Kingma and Ba}{2014}]%
        {Kingma2014AdamAM}
\bibfield{author}{\bibinfo{person}{D.~P. Kingma} {and} \bibinfo{person}{J.
  Ba}.} \bibinfo{year}{2014}\natexlab{}.
\newblock \showarticletitle{Adam: A Method for Stochastic Optimization}.
\newblock \bibinfo{journal}{\emph{CoRR}}  \bibinfo{volume}{abs/1412.6980}
  (\bibinfo{year}{2014}).
\newblock


\bibitem[\protect\citeauthoryear{Ledig, Theis, Huszár, Caballero, Cunningham,
  Acosta, Aitken, Tejani, Totz, Wang, and Shi}{Ledig et~al\mbox{.}}{2017}]%
        {2017Ledigsrgan}
\bibfield{author}{\bibinfo{person}{C. Ledig}, \bibinfo{person}{L. Theis},
  \bibinfo{person}{F. Huszár}, \bibinfo{person}{J. Caballero},
  \bibinfo{person}{A. Cunningham}, \bibinfo{person}{A. Acosta},
  \bibinfo{person}{A. Aitken}, \bibinfo{person}{A. Tejani}, \bibinfo{person}{J.
  Totz}, \bibinfo{person}{Z. Wang}, {and} \bibinfo{person}{W. Shi}.}
  \bibinfo{year}{2017}\natexlab{}.
\newblock \showarticletitle{Photo-Realistic Single Image Super-Resolution Using
  a Generative Adversarial Network}. In \bibinfo{booktitle}{\emph{Proceedings
  of the IEEE Conference on Computer Vision and Pattern Recognition}}.
\newblock
\showISSN{1063-6919}


\bibitem[\protect\citeauthoryear{Leung and Newsam}{Leung and Newsam}{2010}]%
        {leung2010proximate}
\bibfield{author}{\bibinfo{person}{D. Leung} {and} \bibinfo{person}{S.
  Newsam}.} \bibinfo{year}{2010}\natexlab{}.
\newblock \showarticletitle{Proximate Sensing: Inferring What-Is-Where From
  Georeferenced Photo Collections}. In \bibinfo{booktitle}{\emph{Proceedings of
  the IEEE Conference on Computer Vision and Pattern Recognition}}.
  \bibinfo{pages}{2955--2962}.
\newblock


\bibitem[\protect\citeauthoryear{Li, Huang, and Luo}{Li et~al\mbox{.}}{2015}]%
        {li2015using}
\bibfield{author}{\bibinfo{person}{Y. Li}, \bibinfo{person}{J. Huang}, {and}
  \bibinfo{person}{J. Luo}.} \bibinfo{year}{2015}\natexlab{}.
\newblock \showarticletitle{Using User Generated Online Photos to Estimate and
  Monitor Air Pollution in Major Cities}. In
  \bibinfo{booktitle}{\emph{Proceedings of the 7th International Conference on
  Internet Multimedia Computing and Service}}. \bibinfo{pages}{79}.
\newblock


\bibitem[\protect\citeauthoryear{Li, Liu, Yang, and Yang}{Li
  et~al\mbox{.}}{2017}]%
        {li2017face}
\bibfield{author}{\bibinfo{person}{Y. Li}, \bibinfo{person}{S. Liu},
  \bibinfo{person}{J. Yang}, {and} \bibinfo{person}{M. Yang}.}
  \bibinfo{year}{2017}\natexlab{}.
\newblock \showarticletitle{Generative Face Completion}. In
  \bibinfo{booktitle}{\emph{Proceedings of the IEEE Conference on Computer
  Vision and Pattern Recognition}}.
\newblock


\bibitem[\protect\citeauthoryear{Liu, He, Yao, Zhang, Liang, Wang, and
  Hong}{Liu et~al\mbox{.}}{2017}]%
        {liu2017ijgis_landuse}
\bibfield{author}{\bibinfo{person}{X. Liu}, \bibinfo{person}{J. He},
  \bibinfo{person}{Y. Yao}, \bibinfo{person}{J. Zhang}, \bibinfo{person}{H.
  Liang}, \bibinfo{person}{H. Wang}, {and} \bibinfo{person}{Y. Hong}.}
  \bibinfo{year}{2017}\natexlab{}.
\newblock \showarticletitle{Classifying Urban Land Use by Integrating Remote
  Sensing and Social Media Data}.
\newblock \bibinfo{journal}{\emph{International Journal of Geographical
  Information Science}} \bibinfo{volume}{31}, \bibinfo{number}{8}
  (\bibinfo{year}{2017}), \bibinfo{pages}{1675--1696}.
\newblock


\bibitem[\protect\citeauthoryear{Ma, Jia, Sun, Schiele, Tuytelaars, and
  Van~Gool}{Ma et~al\mbox{.}}{2017}]%
        {ma2017pose}
\bibfield{author}{\bibinfo{person}{L. Ma}, \bibinfo{person}{X. Jia},
  \bibinfo{person}{Q. Sun}, \bibinfo{person}{B. Schiele}, \bibinfo{person}{T.
  Tuytelaars}, {and} \bibinfo{person}{L. Van~Gool}.}
  \bibinfo{year}{2017}\natexlab{}.
\newblock \showarticletitle{Pose Guided Person Image Generation}. In
  \bibinfo{booktitle}{\emph{Advances in Neural Information Processing Systems
  (NIPS)}}. \bibinfo{pages}{405--415}.
\newblock


\bibitem[\protect\citeauthoryear{{Mao}, {Li}, {Xie}, {Lau}, {Wang}, and
  {Smolley}}{{Mao} et~al\mbox{.}}{2016}]%
        {mao2016lsgan}
\bibfield{author}{\bibinfo{person}{X. {Mao}}, \bibinfo{person}{Q. {Li}},
  \bibinfo{person}{H. {Xie}}, \bibinfo{person}{R.~Y.~K. {Lau}},
  \bibinfo{person}{Z. {Wang}}, {and} \bibinfo{person}{S.~P. {Smolley}}.}
  \bibinfo{year}{2016}\natexlab{}.
\newblock \showarticletitle{{Least Squares Generative Adversarial Networks}}.
\newblock \bibinfo{journal}{\emph{ArXiv e-prints}} (\bibinfo{date}{Nov.}
  \bibinfo{year}{2016}).
\newblock
\showeprint[arxiv]{cs.CV/1611.04076}


\bibitem[\protect\citeauthoryear{Mirza and Osindero}{Mirza and
  Osindero}{2014}]%
        {mirza2014conditional}
\bibfield{author}{\bibinfo{person}{M. Mirza} {and} \bibinfo{person}{S.
  Osindero}.} \bibinfo{year}{2014}\natexlab{}.
\newblock \showarticletitle{Conditional Generative Adversarial Nets}.
\newblock \bibinfo{journal}{\emph{arXiv preprint arXiv:1411.1784}}
  (\bibinfo{year}{2014}).
\newblock


\bibitem[\protect\citeauthoryear{{Mohamed} and {Lakshminarayanan}}{{Mohamed}
  and {Lakshminarayanan}}{2016}]%
        {2016arXivimplicit}
\bibfield{author}{\bibinfo{person}{S. {Mohamed}} {and} \bibinfo{person}{B.
  {Lakshminarayanan}}.} \bibinfo{year}{2016}\natexlab{}.
\newblock \showarticletitle{{Learning in Implicit Generative Models}}.
\newblock \bibinfo{journal}{\emph{ArXiv e-prints}} (\bibinfo{date}{Oct.}
  \bibinfo{year}{2016}).
\newblock
\showeprint[arxiv]{stat.ML/1610.03483}


\bibitem[\protect\citeauthoryear{Radford, Metz, and Chintala}{Radford
  et~al\mbox{.}}{2016}]%
        {radford2015unsupervised}
\bibfield{author}{\bibinfo{person}{A. Radford}, \bibinfo{person}{L. Metz},
  {and} \bibinfo{person}{S. Chintala}.} \bibinfo{year}{2016}\natexlab{}.
\newblock \showarticletitle{Unsupervised representation learning with deep
  convolutional generative adversarial networks}. In
  \bibinfo{booktitle}{\emph{International Conference on Representation Learning
  (ICRL)}}.
\newblock


\bibitem[\protect\citeauthoryear{Reed, Akata, Yan, Logeswaran, Schiele, and
  Lee}{Reed et~al\mbox{.}}{2016}]%
        {pmlr-v48-reed16}
\bibfield{author}{\bibinfo{person}{S. Reed}, \bibinfo{person}{Z. Akata},
  \bibinfo{person}{X. Yan}, \bibinfo{person}{L. Logeswaran},
  \bibinfo{person}{B. Schiele}, {and} \bibinfo{person}{H. Lee}.}
  \bibinfo{year}{2016}\natexlab{}.
\newblock \showarticletitle{Generative Adversarial Text to Image Synthesis}. In
  \bibinfo{booktitle}{\emph{Proceedings of the 33rd International Conference on
  Machine Learning (ICML)}} \emph{(\bibinfo{series}{Proceedings of Machine
  Learning Research})}, \bibfield{editor}{\bibinfo{person}{Maria~Florina
  Balcan} {and} \bibinfo{person}{Kilian~Q. Weinberger}} (Eds.),
  Vol.~\bibinfo{volume}{48}. \bibinfo{publisher}{PMLR}, \bibinfo{address}{New
  York, New York, USA}, \bibinfo{pages}{1060--1069}.
\newblock


\bibitem[\protect\citeauthoryear{Regmi and Borji}{Regmi and Borji}{2018a}]%
        {Regmi_2018_CVPR}
\bibfield{author}{\bibinfo{person}{K. Regmi} {and} \bibinfo{person}{A. Borji}.}
  \bibinfo{year}{2018}\natexlab{a}.
\newblock \showarticletitle{Cross-View Image Synthesis Using Conditional GANs}.
  In \bibinfo{booktitle}{\emph{Proceedings of the IEEE Conference on Computer
  Vision and Pattern Recognition}}.
\newblock


\bibitem[\protect\citeauthoryear{Regmi and Borji}{Regmi and Borji}{2018b}]%
        {2018arXivcrossview}
\bibfield{author}{\bibinfo{person}{K. Regmi} {and} \bibinfo{person}{A. Borji}.}
  \bibinfo{year}{2018}\natexlab{b}.
\newblock \showarticletitle{{Cross-view image synthesis using geometry-guided
  conditional GANs}}.
\newblock \bibinfo{journal}{\emph{ArXiv e-prints}} (\bibinfo{date}{Aug.}
  \bibinfo{year}{2018}).
\newblock
\showeprint[arxiv]{cs.CV/1808.05469}


\bibitem[\protect\citeauthoryear{Salimans, Goodfellow, Zaremba, Cheung,
  Radford, and Chen}{Salimans et~al\mbox{.}}{2016}]%
        {salimans2016improved}
\bibfield{author}{\bibinfo{person}{T. Salimans}, \bibinfo{person}{I.
  Goodfellow}, \bibinfo{person}{W. Zaremba}, \bibinfo{person}{V.i Cheung},
  \bibinfo{person}{A. Radford}, {and} \bibinfo{person}{X. Chen}.}
  \bibinfo{year}{2016}\natexlab{}.
\newblock \showarticletitle{Improved Techniques for Training GANs}. In
  \bibinfo{booktitle}{\emph{Advances in Neural Information Processing Systems
  (NIPS)}}. \bibinfo{pages}{2234--2242}.
\newblock


\bibitem[\protect\citeauthoryear{{Tran}, {Yin}, and {Liu}}{{Tran}
  et~al\mbox{.}}{2017}]%
        {2017arXivface}
\bibfield{author}{\bibinfo{person}{L. {Tran}}, \bibinfo{person}{X. {Yin}},
  {and} \bibinfo{person}{X. {Liu}}.} \bibinfo{year}{2017}\natexlab{}.
\newblock \showarticletitle{{Representation Learning by Rotating Your Faces}}.
\newblock \bibinfo{journal}{\emph{ArXiv e-prints}} (\bibinfo{date}{May}
  \bibinfo{year}{2017}).
\newblock
\showeprint[arxiv]{cs.CV/1705.11136}


\bibitem[\protect\citeauthoryear{Wang, Korayem, Blanco, and Crandall}{Wang
  et~al\mbox{.}}{2016}]%
        {wang2016tracking}
\bibfield{author}{\bibinfo{person}{J. Wang}, \bibinfo{person}{M. Korayem},
  \bibinfo{person}{S. Blanco}, {and} \bibinfo{person}{D. Crandall}.}
  \bibinfo{year}{2016}\natexlab{}.
\newblock \showarticletitle{Tracking Natural Events Through Social Media and
  Computer Vision}. In \bibinfo{booktitle}{\emph{2016 ACM on Multimedia
  Conference}}. \bibinfo{pages}{1097--1101}.
\newblock


\bibitem[\protect\citeauthoryear{Workman, Zhai, Crandall, and Jacobs}{Workman
  et~al\mbox{.}}{2017}]%
        {Workman2017AUM}
\bibfield{author}{\bibinfo{person}{S. Workman}, \bibinfo{person}{M. Zhai},
  \bibinfo{person}{D.~J. Crandall}, {and} \bibinfo{person}{N. Jacobs}.}
  \bibinfo{year}{2017}\natexlab{}.
\newblock \showarticletitle{A Unified Model for Near and Remote Sensing}. In
  \bibinfo{booktitle}{\emph{Proceedings of the IEEE International Conference on
  Computer Vision}}. \bibinfo{pages}{2707--2716}.
\newblock


\bibitem[\protect\citeauthoryear{Yeh$^\ast$, Chen$^\ast$, Lim, Schwing,
  Hasegawa-Johnson, and Do}{Yeh$^\ast$ et~al\mbox{.}}{2017}]%
        {yeh2017semantic}
\bibfield{author}{\bibinfo{person}{R.~A. Yeh$^\ast$}, \bibinfo{person}{C.
  Chen$^\ast$}, \bibinfo{person}{T. Lim}, \bibinfo{person}{A.~G. Schwing},
  \bibinfo{person}{M. Hasegawa-Johnson}, {and} \bibinfo{person}{M.~N. Do}.}
  \bibinfo{year}{2017}\natexlab{}.
\newblock \showarticletitle{Semantic Image Inpainting with Deep Generative
  Models}. In \bibinfo{booktitle}{\emph{Proceedings of the IEEE Conference on
  Computer Vision and Pattern Recognition}}.
\newblock
\newblock
\shownote{$^\ast$ equal contribution.}


\bibitem[\protect\citeauthoryear{Zhai, Bessinger, Workman, and Jacobs}{Zhai
  et~al\mbox{.}}{2017}]%
        {zhai2017cvpr}
\bibfield{author}{\bibinfo{person}{M. Zhai}, \bibinfo{person}{Z. Bessinger},
  \bibinfo{person}{S. Workman}, {and} \bibinfo{person}{N. Jacobs}.}
  \bibinfo{year}{2017}\natexlab{}.
\newblock \showarticletitle{Predicting Ground-Level Scene Layout from Aerial
  Imagery}. In \bibinfo{booktitle}{\emph{Proceedings of the IEEE Conference on
  Computer Vision and Pattern Recognition}}.
\newblock


\bibitem[\protect\citeauthoryear{Zhang, Xu, Li, Zhang, Huang, Wang, and
  Metaxas}{Zhang et~al\mbox{.}}{2017}]%
        {zhang2017stackgan}
\bibfield{author}{\bibinfo{person}{H. Zhang}, \bibinfo{person}{T. Xu},
  \bibinfo{person}{H. Li}, \bibinfo{person}{S. Zhang}, \bibinfo{person}{X.
  Huang}, \bibinfo{person}{X. Wang}, {and} \bibinfo{person}{D. Metaxas}.}
  \bibinfo{year}{2017}\natexlab{}.
\newblock \showarticletitle{Stackgan: Text to Photo-Realistic Image Synthesis
  with Stacked Generative Adversarial Networks}. In
  \bibinfo{booktitle}{\emph{Proceedings of the IEEE International Conference on
  Computer Vision}}. \bibinfo{pages}{5907--5915}.
\newblock


\bibitem[\protect\citeauthoryear{Zhu, Park, Isola, and Efros}{Zhu
  et~al\mbox{.}}{2017b}]%
        {CycleGAN2017}
\bibfield{author}{\bibinfo{person}{J. Zhu}, \bibinfo{person}{T. Park},
  \bibinfo{person}{P. Isola}, {and} \bibinfo{person}{A.~A Efros}.}
  \bibinfo{year}{2017}\natexlab{b}.
\newblock \showarticletitle{Unpaired Image-to-Image Translation using
  Cycle-Consistent Adversarial Networks}. In
  \bibinfo{booktitle}{\emph{Proceedings of the IEEE International Conference on
  Computer Vision}}.
\newblock


\bibitem[\protect\citeauthoryear{{Zhu}, {Deng}, and {Newsam}}{{Zhu}
  et~al\mbox{.}}{2018}]%
        {zhu2018arxiv}
\bibfield{author}{\bibinfo{person}{Y. {Zhu}}, \bibinfo{person}{X. {Deng}},
  {and} \bibinfo{person}{S. {Newsam}}.} \bibinfo{year}{2018}\natexlab{}.
\newblock \showarticletitle{{Fine-Grained Land Use Classification at the City
  Scale Using Ground-Level Images}}.
\newblock \bibinfo{journal}{\emph{ArXiv e-prints}} (\bibinfo{date}{Feb.}
  \bibinfo{year}{2018}).
\newblock
\showeprint[arxiv]{cs.CV/1802.02668}


\bibitem[\protect\citeauthoryear{Zhu, Liu, and Newsam}{Zhu
  et~al\mbox{.}}{2017a}]%
        {Zhu2017Activity}
\bibfield{author}{\bibinfo{person}{Y. Zhu}, \bibinfo{person}{S. Liu}, {and}
  \bibinfo{person}{S. Newsam}.} \bibinfo{year}{2017}\natexlab{a}.
\newblock \showarticletitle{Large-Scale Mapping of Human Activity Using
  Geo-Tagged Videos}. In \bibinfo{booktitle}{\emph{Proceedings of the 25th ACM
  SIGSPATIAL International Conference on Advances in Geographic Information
  Systems}}. \bibinfo{pages}{68:1--68:4}.
\newblock


\bibitem[\protect\citeauthoryear{Zhu and Newsam}{Zhu and Newsam}{2015}]%
        {zhu2015land}
\bibfield{author}{\bibinfo{person}{Y. Zhu} {and} \bibinfo{person}{S. Newsam}.}
  \bibinfo{year}{2015}\natexlab{}.
\newblock \showarticletitle{Land Use Classification Using Convolutional Neural
  Networks Applied to Ground-Level Images}. In
  \bibinfo{booktitle}{\emph{Proceedings of the 23rd SIGSPATIAL International
  Conference on Advances in Geographic Information Systems}}.
  \bibinfo{pages}{61}.
\newblock


\bibitem[\protect\citeauthoryear{Zhu and Newsam}{Zhu and Newsam}{2016}]%
        {zhu2016spatio}
\bibfield{author}{\bibinfo{person}{Y. Zhu} {and} \bibinfo{person}{S. Newsam}.}
  \bibinfo{year}{2016}\natexlab{}.
\newblock \showarticletitle{Spatio-Temporal Sentiment Hotspot Detection Using
  Geotagged Photos}. In \bibinfo{booktitle}{\emph{Proceedings of the 24th ACM
  SIGSPATIAL International Conference on Advances in Geographic Information
  Systems}}. \bibinfo{pages}{76}.
\newblock


\end{thebibliography}
